\pdfoutput=1

\documentclass[11pt]{article}

\usepackage[final]{acl}

\usepackage{times}
\usepackage{latexsym}

\usepackage[T1]{fontenc}

\usepackage[utf8]{inputenc}

\usepackage{microtype}

\usepackage{inconsolata}

\usepackage{graphicx}
\usepackage{booktabs}
\usepackage{subcaption}
\usepackage{enumitem}
\usepackage{cleveref} 

\title{Is analogy enough to draw novel adjective-noun inferences?}

\author{
 \textbf{Hayley Ross,\textsuperscript{1}}
 \textbf{Kathryn Davidson,\textsuperscript{1}}
 \textbf{Najoung Kim\textsuperscript{2}}
\\
 \textsuperscript{1}Harvard University
 \textsuperscript{2}Boston University
\\
 \small{
   \texttt{hayleyross@g.harvard.edu} \quad 
   \texttt{kathryndavidson@fas.harvard.edu} \quad
   \texttt{najoung@bu.edu}
 }
}

\begin{document}
\maketitle
\begin{abstract}
Recent work \citep{ross_fake_2025, ross_is_2024} has argued that the ability of humans and LLMs respectively to generalize to novel adjective-noun combinations shows that they each have access to a compositional mechanism to determine the phrase's meaning and derive inferences. 
We study whether these inferences can instead be derived by analogy to known inferences, without need for composition. 
We investigate this by (1) building a model of analogical reasoning using similarity over lexical items, and (2) asking human participants to reason by analogy.
While we find that this strategy works well for a large proportion of the dataset of \citet{ross_fake_2025}, there are novel combinations for which both humans and LLMs derive convergent inferences but which are not well handled by analogy. 
We thus conclude that the mechanism humans and LLMs use to generalize in these cases cannot be fully reduced to analogy, and likely involves composition.
\end{abstract}

\section{Introduction}

How are humans able to generalize to complex linguistic expressions they have not encountered before?
One view on how this can be achieved is through a mechanism of composition, determining the meaning of the phrase and any resulting inferences from the meanings of its parts (\citealp{partee_formal_2009, szabo_case_2012}, i.a.).
Others, however, believe that composition is not required: mechanisms such as analogy are sufficient to explain humans' ability to generalize to novel phrases (\citealp{bybee_language_2010, ambridge_against_2020} i.a.). 
The same question arises when we study LLMs' ability to generalize. If they can generalize to novel phrases, is this evidence that they must be composing these phrases from their subparts, or is there another way to achieve the same results?

\begin{figure}[t]
  \includegraphics[width=\columnwidth]{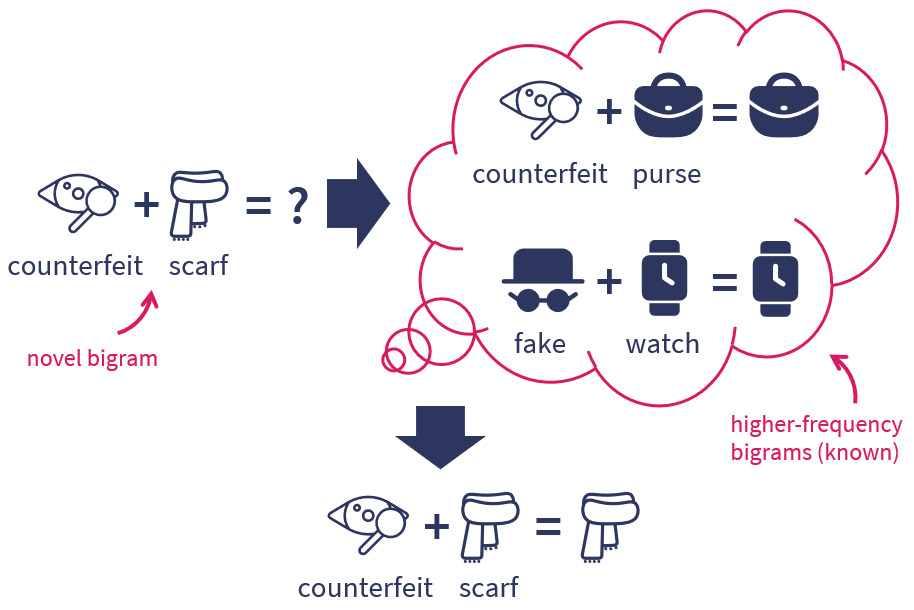}
  \caption{Possible analogical reasoning to infer that \textit{counterfeit scarf} is a scarf, since a \textit{counterfeit purse} is a purse and a \textit{fake} (or \textit{counterfeit}) \textit{watch} is a watch.}
  \label{fig:model-diagram}
\end{figure}

\citet{ross_fake_2025} argue that humans must be using composition, since they converge on the inferences of at least some combinations that they are assumed never to have seen before (e.g., for \emph{fake reef} or \emph{counterfeit scarf}, which never appear in a large corpus). \citet{ross_is_2024} suggest a similar conclusion for LLMs based on the same dataset, since LLMs show reasonably human-like behavior on at least some bigrams that are assumed not to be in the LLMs' training datasets. 
These combinations are interesting because the membership inferences targeted (e.g.,~``Is a counterfeit scarf still a scarf?'') depend not just on the adjective but also on the noun, involving significant detail about how exactly the adjective affects the noun and what properties are important for membership in that noun category in typical situations.

This paper questions these conclusions, and investigates whether this task can in fact be solved by analogical reasoning, without composition.
For example, for \textit{counterfeit scarf}, one might reason (as in Figure \ref{fig:model-diagram}): ``Is a counterfeit scarf still a scarf? 
A scarf is an accessory like a watch or a purse, and a counterfeit watch is still a watch, and a counterfeit purse is still a purse, so a counterfeit scarf is most likely still a scarf''. 
This skips the compositional step of combining the meanings of the words to derive the meaning of the bigram and further violates the principle of compositionality as stated by \citet{szabo_case_2012} by referring to information beyond the meaning of the bigram's parts, namely the inferences associated with other adjective-noun bigrams.

We investigate analogical reasoning through two complementary approaches. 
First, we build a computational model of analogical reasoning which 
attempts to derive ratings for the low-frequency and zero-frequency (assumed novel) bigrams in the dataset of \citet{ross_fake_2025}, by analogy to the high-frequency ones. 
A computational model allows us to precisely define what we mean by analogy, and explore the consequences of different implementation decisions.
Second, we ask human participants to reason analogically, guided by examples and their own intuition of what analogy means. We then evaluate how often they can produce an analogy, and whether the resulting rating distributions derived analogically are the same as the distributions from \citet{ross_fake_2025}, where no instructions on how to reason were given. We find that the ratings derived by analogy significantly differ for several bigrams, suggesting that the original participants did not derive (all) their ratings by analogy. 

Between the two methods, we find convincing evidence that while analogical reasoning produces similar results in many cases, it is not sufficient to derive the full set of inference data. 
Thus, we find support for the view that humans must have access to a compositional mechanism.
Further, our analogy model performs worse on novel bigrams than the best LLM in \citet{ross_is_2024}, and our analogy model's successes and failures correlate poorly with those of the best LLM. This suggests that the LLM is not (just) using analogy in the cases where it can generalize, 
and supports the claim in \citet{ross_is_2024} that such LLMs are performing some kind of composition (productively combining the meaning of adjective and noun) in these cases. We share our code and data on GitHub.\footnote{\url{https://github.com/rossh2/artificial-intelligence/}}

\section{Related Work}

So-called ``privative'' adjectives such as \textit{fake} pose a challenge for compositional accounts of semantics, since they cannot be simply intersected with the noun \citep{kamp_prototype_1995}. Multiple accounts have been proposed for how composition with privative adjectives should work (\citealp{partee_privative_2010,del_pinal_dual_2015,martin_compositional_2022,guerrini_keeping_2024} i.a.). 

Most previous computational work on adjective-noun composition using distributional semantics does not discuss privative adjectives \citep{baroni_nouns_2010,vecchi_spicy_2017,hartung_learning_2017}. \citet{boleda_intensionality_2013} cover 16 ``non-intensional'' adjectives, including two which are commonly taken to be privative (\textit{former}, \textit{mock}; see \citet{nayak_dictionary_2014} for a classification). \citeauthor{boleda_intensionality_2013} build distributional semantic models of adjective-noun composition that use vector addition and matrix multiplication to model adjective-noun composition, but they do not cover analogy. \citet{cappelle_facing_2018} study the distributional semantics of \textit{fake} and bigrams in which it occurs, but do not implement any method of composition or generalization.

\citet{ross_fake_2025} gather a large quantity of offline human judgments on (privative) adjectives and their membership inferences, discussed further in Section \ref{sec:elm-dataset}, and \citet{ross_is_2024} extend this dataset to assess LLMs. 
While \citet{ross_is_2024} do propose a simple analogy baseline to compare to their LLMs, we propose an improved, more powerful and configurable analogy model and present a detailed analysis of its performance.

Analogy has been much studied as a core component of human reasoning (see \citealp{hofstadter_epilogue_2001} for an overview), and approaches such as construction grammar propose that analogy to known exemplars can be used to understand any novel phrase \citep{bybee_language_2010,ambridge_against_2020}. \citet{rambelli_compositionality_2024} propose a computational model of this process based on distributional semantics. While we also build our computational model around analogy between phrases, we only attempt to derive membership inferences from the analogy, and avoid commitment to whether the full meaning of the phrase can be accessed by analogy. 

\section{Human Judgment Dataset} \label{sec:elm-dataset}

\citet{ross_fake_2025} present a dataset of human judgments on adjective-noun inferences of the form ``Is an \{adjective\} \{noun\} still a \{noun\}?'' on a 5-point Likert scale. 
The dataset covers 798 bigrams (102 nouns crossed with 6 typically privative and 6 typically subsective adjectives, filtered to only include combinations that make sense).\footnote{In this paper, we follow \citet{ross_fake_2025} in using ``(typically-) privative / subsective adjective'' to refer to adjectives historically classified as such, which often but not always result in the respective inference.} 
In this dataset, the question is presented out of the blue as a generic, rather than in a discourse context. The additional information in a discourse can sometimes determine the inference on its own (without needing to interpret the bigram at all), whereas the out of the blue setting requires some kind of reasoning strategy (composition, analogy or otherwise) to determine the inference. 
180 of the 798 bigrams are zero frequency in the C4 pretraining corpus \citep{raffel_exploring_2020}, which \citet{ross_is_2024} take as a proxy for the undisclosed pretraining corpora of the models they study. These bigrams are assumed to be novel to both humans and LLMs.
A bigram is referred to as high frequency if it is in the top quartile of bigrams they study. 

\citet{ross_fake_2025} show that the membership inference in question depends on both the adjective and the noun, with bigrams with ``subsective'' adjectives usually yielding subsective inferences (e.g., ``a homemade N is an N'', but not always: consider \textit{homemade cat}), while bigrams with ``privative'' adjectives such as \textit{fake crowd} elicit a wide distribution of ratings from subsective (``is'') to privative (``is not''), with high variance for many (but not all) bigrams.
Varying ratings between participants are expected in this setting, since we are dealing not only with the lexicon but also with a broad question (a linguistic generic) which may depend on participants' world knowledge. 
Participants nonetheless show convergent ratings for many zero-frequency bigrams, demonstrating their ability to generalize and implying a shared underlying mechanism.

\section{Analogy Model} \label{sec:analogy-model}

\subsection{Algorithm} \label{sec:analogy-model-algorithm}

We implement a computational model of analogy which 
is ``trained'' on the human ratings from \citet{ross_fake_2025} for a set of common (high-frequency) bigrams, which are stored in the model's memory.  
This is intended to imitate human prior experience with certain bigrams, where they may have learned that, for instance, a \textit{counterfeit watch} is still a \textit{watch}. 
Humans are known to store frequent multi-word expressions even when those expressions are compositional, not just when they are idiomatic (\citealp{arnon_more_2010,tremblay_holistic_2010,caldwell-harris_measuring_2012}, i.a.), so it is plausible to assume that they can also store the associated inferences.
Specifically, we consider the top quartile of bigrams in \citet{ross_fake_2025} as ``known'', i.e.,~in the training set. (Appendix \ref{sec:hyperparameters} also explores an alternative approach where the training set is balanced evenly across adjectives.)

\begin{figure}
    \centering
    \includegraphics[width=\columnwidth]{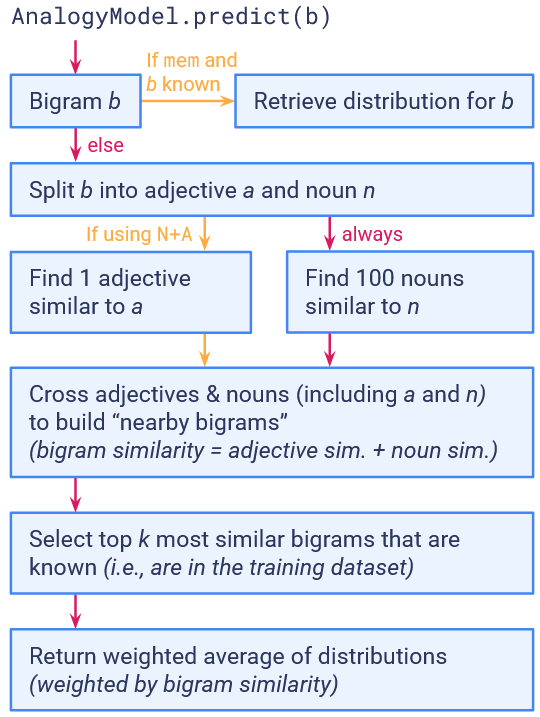}
    \caption{Algorithm for the analogy model. Yellow paths are dependent on the configuration options \texttt{mem} and \texttt{N+A} (Noun + Adjective). $k$ is a hyperparameter.}
    \label{fig:analogy-model-algorithm}
\end{figure}

Given these known bigrams, the model predicts the ratings for the remainder of the bigrams by analogy to similar bigrams in its training set, via the algorithm in Figure~\ref{fig:analogy-model-algorithm}. The setting \texttt{mem} configures whether this algorithm is also applied to bigrams in the training set, as if they were not known; we discuss in Section \ref{sec:am-human-discussion} what is more human-like.

The model stores and predicts the entire rating distribution for each bigram, rather than a single rating. 
As \citet{ross_is_2024} discuss in the context of LLMs, it is not clear how to evaluate the alignment of a single rating against high variance distributions like the human data we are taking as the evaluation target. 
As discussed in Section \ref{sec:elm-dataset}, such high variation is a natural consequence of working with the lexicon, but does
necessitate a more complex metric than just accuracy to assess model fit. 
We use same metric that \citet{ross_is_2024} use for LLMs: the Jensen-Shannon divergence between the model-predicted rating distribution and the human rating distribution for each bigram. We compute an aggregate score by averaging across all bigrams. 
We report this aggregate score as well as the average score over zero-frequency bigrams (presumed to be novel to both humans and LLMs) to measure its ability to generalize. These zero-frequency bigrams are always held out from the model.

\begin{figure*}[t]
  \includegraphics[width=\textwidth]{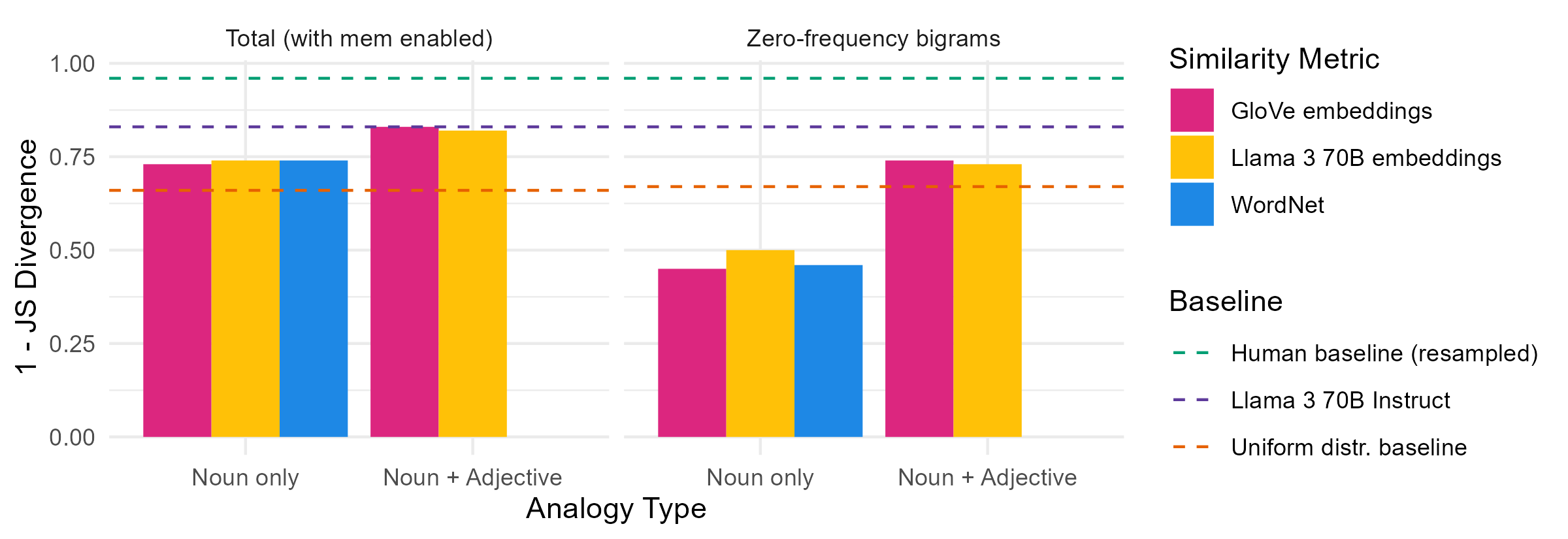}
  \caption{Average JS divergence between distributions produced by the analogy model and human distributions from \citet{ross_fake_2025} on zero-frequency bigrams and on the whole dataset (with memorization of the training set). Additional results are given in Table \ref{tab:analogy-model-results} in the Appendix.}
  \label{fig:analogy-model-barplot}
\end{figure*}

Implementing analogical reasoning in a computational model allows us to define precisely what we mean by analogy and test the effects of these implementation choices.
We explore two types of analogy: either just over nouns (\textit{counterfeit scarf} $\rightarrow$ \textit{counterfeit watch}),\footnote{We see in Section \ref{sec:human-experiment} that this is a popular human strategy: humans choose an analogy over just nouns 58\% of the time.} or allowing analogy over both noun and up to one additional adjective (\textit{counterfeit scarf} $\rightarrow$ \textit{fake watch}; \texttt{N+A} setting). 
We allow the model to retain $k \leq 5$ nearby bigrams (after filtering to bigrams in the training set) to impose constraints akin to human working memory \citep{cowan_magical_2001,adam_clear_2017}. The exact value of $k$ is a hyperparameter optimized on the training set (with memorization disabled).
Appendix \ref{sec:hyperparameters} also discusses the case where $k=1$, i.e.~where the model only considers the most similar bigram, which is a plausible route for humans. 

We calculate word similarity in three ways: (1) cosine similarity over GloVe embeddings \citep{pennington_glove_2014}; (2) cosine similarity over embeddings from Llama 3 70B Instruct \citep{dubey_llama_2024} and (3) Wu-Palmer similarity over the WordNet taxonomy \citep{wu_verb_1994,miller_wordnet_1995}. 
Llama 3 70B Instruct was selected as the source for LLM embeddings because this was the model with the highest performance in \citet{ross_is_2024}. 
To derive word embeddings from Llama, we pass each word individually to the LLM and average the hidden states of the subword tokens in the final layer.\footnote{We could alternatively pool the embeddings from the initial embedding layer, but the absence of contextualization in this approach may degrade results for multi-token words (\char`~ 40\% of our dataset).  
Nevertheless, we show in Appendix \ref{sec:hyperparameters} that results are similar in this setting.} 
Wu-Palmer similarity groups nouns\footnote{Strictly, the metric groups noun synsets (``senses''); we use the 2 most common synsets per noun.} that share common hypernyms in WordNet, penalized by how broad that hypernym is. 
Using WordNet allows us to measure similarity based solely on a human-created dataset, as opposed to distributionally derived embeddings.
Since WordNet does not provide a taxonomy of adjectives, this approach is limited to noun-only analogies.

\subsection{Results}

Figure \ref{fig:analogy-model-barplot} shows the performance of the different analogy model configurations on the whole dataset (allowing memorization of the training set) and on held-out, zero-frequency bigrams (assumed to be novel to humans and LLMs). More details, including results for privative adjectives only and for single-bigram analogies ($k=1$), are given in Appendix \ref{sec:hyperparameters} (Table \ref{tab:analogy-model-results}).

\paragraph{GloVe embeddings.}

Both the noun-only and N+A setting perform well overall, with the N+A setting appearing to be on par with LLM performance. However, we find that this is reliant on memorizing the training set; neither setting generalizes well to zero-frequency bigrams. In particular, noun-only analogies perform below a uniform distribution baseline on zero-frequency bigrams. 

\paragraph{WordNet.} \label{sec:wordnet-results}

Perhaps surprisingly, we find that this qualitatively different similarity metric yields very similar results to using GloVe embeddings, at least in the noun-only case where this metric is defined. We discuss the implication further in Section \ref{sec:model-inadequacy}.

\paragraph{Llama Embeddings.} \label{sec:llama-embedding-results}

Using the embeddings derived from Llama 3 70B Instruct also does not improve performance significantly compared to using GloVe, though we see a small increase for the noun-only setting---see also the discussion in Section \ref{sec:model-inadequacy}. 

\paragraph{Error Analysis.}

To investigate where the analogy model fails, we fit a linear regression in R \citep{r_core_team_r_2023} that predicts the JS divergence of the best-performing model from the adjective class (subsective vs. privative), human rating mean and human rating SD, with an interaction between adjective class and mean. Including the human SD allows us to target bigrams with divergent ratings; including an interaction of adjective class and mean allows us to pick out e.g.~bigrams with subsective adjectives but privative ratings.

All main effects and the interaction are significant: JS divergence is lower for privative-class adjectives, higher for bigrams with subsective-class adjectives with privative ratings (i.e., low mean ratings, such as \textit{homemade money} or \textit{tiny abundance}), higher for privative-class bigrams with subsective ratings (i.e., high mean ratings, such as \textit{false rumor} or \textit{counterfeit watch}), and lower for bigrams with a high human standard deviation. 
The fact that it struggles on bigrams like \textit{homemade money} ($\textrm{JS} = 0.81$) and \textit{tiny abundance} ($\textrm{JS} = 0.58$) in particular is not surprising, given that these adjectives are subsective for all except two bigrams in the model's pool of analogy candidates. 

\subsection{Discussion: Effect of Similarity Metric} \label{sec:model-inadequacy}
 
The similarity metric used is not a main modulator of model performance.
One possible explanation is that the analogies found by our model may often be suboptimal or inadequate, regardless of the similarity metric used. There are two potential sources of this inadequacy: first, analogical reasoning may inherently be a flawed approach for some bigrams. Second, the training set may be so sparse that 
the model cannot retrieve sufficiently similar nouns or bigrams to adequately support analogical reasoning. 
After all, our training set contains ratings for only 279 bigrams using 89 nouns (of 102 nouns in the original dataset).\footnote{The 102 nouns were selected by \citeauthor{ross_fake_2025} such that each noun has at least one closely related other noun.}
While we cannot fully tease these two possibilities apart with our current experiments, Appendix \ref{sec:am-extra-data} explores adding data from the human rating experiment in Section \ref{sec:human-experiment}. 

\subsection{Discussion: Humans} \label{sec:am-human-discussion}

Working with lexical semantics requires us to deal with per-bigram distributions and a distribution comparison metric, rather than proportions of correct answers or significant effects in a regression. This makes interpretation of the results more complicated. It is not clear at what threshold to conclude that the model captures human performance, versus what amount of JS divergence represents noise/artifacts generated by the relatively small distribution sample size in the human experiment ($n=12$ per bigram).
Short of replicating the human experiment in \citet{ross_fake_2025} and calculating the JS divergence between the two, we have three points of reference:
(1) We can approximate a human JS divergence by resampling from the human distribution. This yields an average JS divergence of just 0.05;
(2) The best LLM performance that achieves JS divergence of 0.17 both overall and on zero-frequency bigrams \citep{ross_is_2024};
(3) The ratings collected from the experiment in Section \ref{sec:human-experiment}, where humans are asked to perform the same task as the analogy model, yield an overall JS divergence of 0.16 compared to the original distributions.

Our analogy model achieves a JS divergence of 0.17 at best, when allowed to memorize its training data; 0.25 when it does not memorize it. On zero-frequency bigrams, the best score is 0.25. 
While the results are impressive with memorization, its ability to generalize to zero-frequency bigrams is 8 points worse than LLMs and 11 points worse than humans.
This suggests that our analogy model does not fully capture human behavior.
While a key part of the modeling assumption is that the training data represents humans' known and memorized bigrams, it is still unclear whether it is human-like to return the exact perfect distribution---all the more so considering that we typically ask humans to give single ratings, not entire distributions.

As an alternative metric, we conduct per-bigram Kolmogorov-Smirnoff tests (Holm-Bonferroni adjusted) comparing the distributions predicted by the analogy model to the human distributions. We find that with memorization of the training set, 10 of the predicted distributions are significantly different ($p < 0.05$), of which 3 are zero-frequency bigrams; without memorization, this rises to 20. Since we only have a sample size of $n=12$, this is a conservative estimate. 
Figure \ref{fig:analogy-model-different-dists} in Appendix \ref{sec:am-signif-different} shows a selection of such distributions.
The fact that the analogy model significantly deviates from the correct distribution for these cases supports our conclusion that while analogy is successful in most cases, it does not offer a full explanation.

\subsection{Discussion: LLMs} \label{sec:am-llms}

It may seem striking that the analogy model can achieve the same overall JS divergence as Llama 3 70B Instruct, the best model studied by \citet{ross_is_2024}, when we allow training set memorization. However, comparing results on the zero-frequency bigrams (and also on performance without \texttt{mem}, see Table \ref{tab:analogy-model-results}) shows that Llama 3 70B Instruct generalizes much better than our analogy model. 
Further, fitting a linear regression to predict the LLM's JS divergence per-bigram from the Llama embedding analogy model's divergence shows that although the effect is significant ($p < 0.001$), this only explains 12\% of the variance in the LLM's ratings ($R^2 = 0.12$; $R^2=0.04$ with \texttt{mem} enabled). In other words, the LLM's behavior is not particularly well explained by the analogy model, and it does not succeed and fail in the same places. 

\begin{figure*}[t]
\centering
\begin{subfigure}[t]{.45\linewidth}
    \centering
    \includegraphics[width=\linewidth]{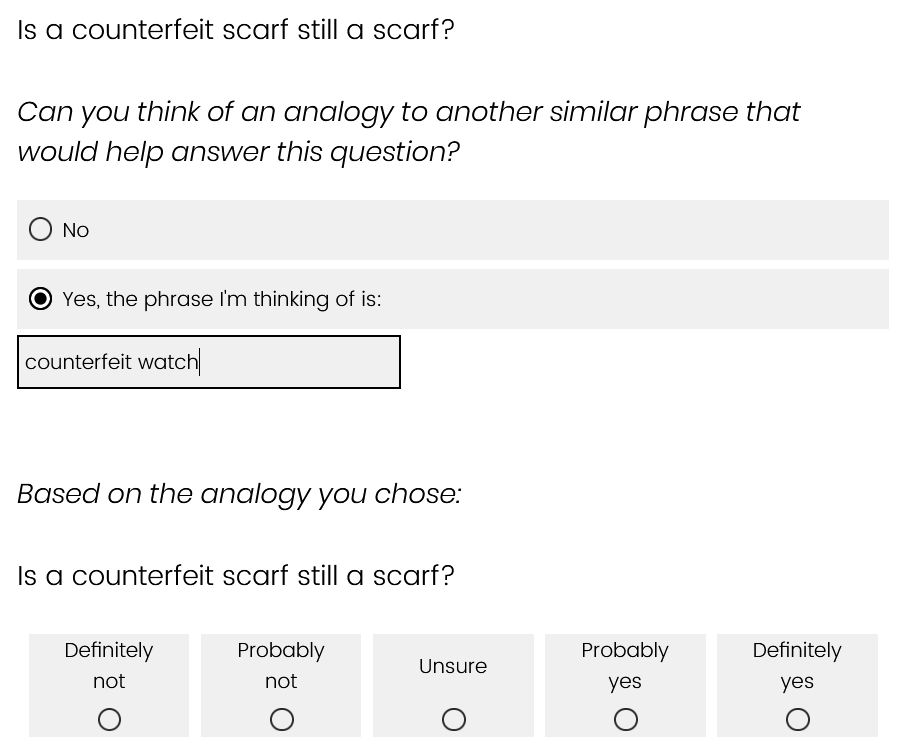}
    \caption{Path when analogy found.
    \label{fig:analogy-path-screenshot}}
  \end{subfigure}
  \begin{subfigure}[t]{.45\linewidth}
    \centering
    \includegraphics[width=\linewidth]{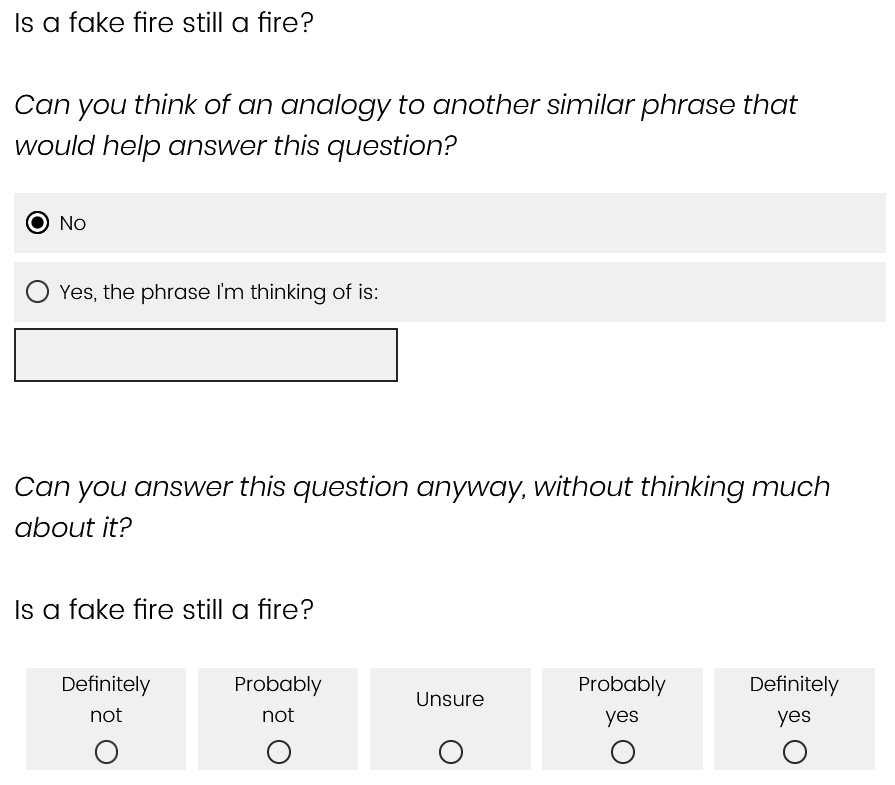}
    \caption{Path when no analogy found.
    \label{fig:no-analogy-path-screenshot}}
  \end{subfigure}
  \caption{Screenshots of questions in the analogy prompting experiment.} \label{fig:qualtrics-screenshots}
\end{figure*}

\section{Human Analogical Reasoning} \label{sec:human-experiment}

While the analogy model allows us to precisely control the mechanism and data used for analogical reasoning, it also suffers from an artificial restriction on the bigrams to which it can draw an analogy: its training dataset is strictly limited to the bigrams that \citet{ross_fake_2025} gathered human ratings for. 
Actual human analogical reasoning would not be limited in the same way, and is likely to involve a much wider range of analogy targets. 
In this experiment on human participants, we expand the definition of analogy to whatever our participants construe as analogy (given our instructions and training examples), enabling access to whatever bigrams they are able to come up with as suitable analogies. This allows us to measure two things: (1) how easy it is for people to come up with analogies at all, and (2) what effect analogical reasoning has on the resulting rating distributions.

\subsection{Method} \label{sec:human-exp-method}

We select 96 bigrams from the 798 bigrams from \citet{ross_fake_2025} such that they are evenly balanced by adjective and by zero vs. top quartile frequency, and all have convergent human rating distributions ($\mu \leq 2$ or $\mu \geq 4$ on the 5-point scale).\footnote{We also attempt to include a high proportion of bigrams where analogy might be hard---see Appendix \ref{app:analogy-hard}. For example, we adversarially pick some nouns for \textit{homemade} which are likely to yield privative judgments, such as \textit{homemade money}.} 

For each bigram, we show participants the question ``Is an \{adjective\} \{noun\} still a \{noun\}?'' and first ask them whether they are able to come up an analogy that helps them answer the question. We then ask them to answer the question, either using the analogy or not, depending on their first answer. 
Screenshots of each path are shown in Figure \ref{fig:qualtrics-screenshots}.
Participants first see an explanation of what we mean by analogy, including an example (\textit{toy hippo} $\rightarrow$ \textit{toy elephant}), followed by three training examples 
which include another example of an analogy (\textit{melted plastic} $\rightarrow$ \textit{melted wax/chocolate}). The full instructions, including our description of ``analogy'', are given in Appendix \ref{app:experiment-training}. The analogy text field is limited to 1-3 words to encourage analogy to adjective-noun phrases (pilot participants sometimes typed a reasoning process into the field).

We recruited 176 native American English speakers\footnote{
See Appendix \ref{app:native-speaker} for detailed criteria.} on Prolific, of which we excluded 33 for not meeting our native speaker criteria, failed attention checks, or failing to adequately follow our instructions for analogical reasoning (verified based on manual inspection and regular expression searches on the free text entry fields).

\subsection{Results} \label{sec:human-exp-results}

\begin{figure}[t]
  \includegraphics[width=\columnwidth]{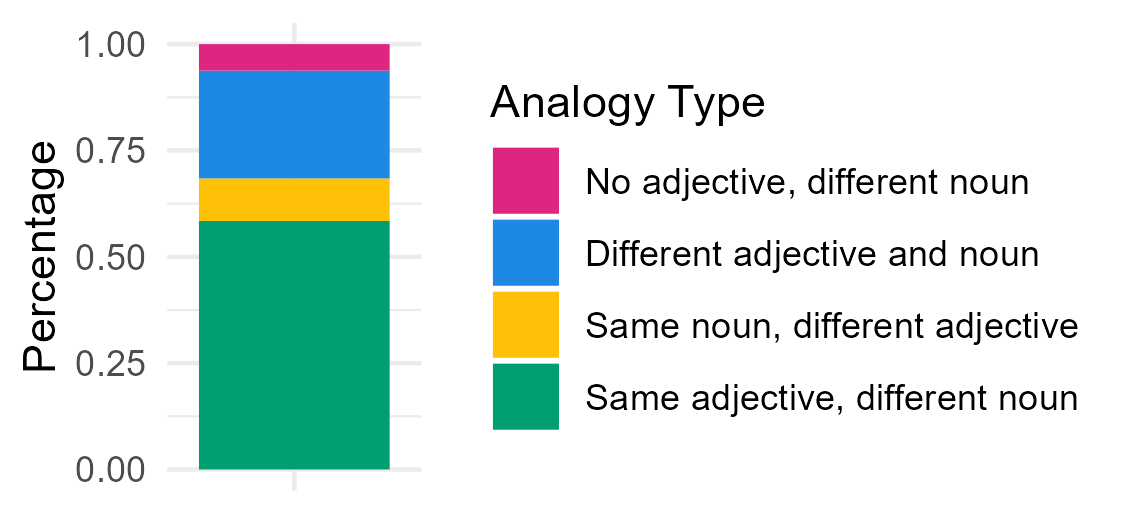}
  \caption{Types of analogy chosen by participants.}
  \label{fig:analogy-types}
\end{figure}

\begin{figure*}[t]
  \includegraphics[width=\textwidth]{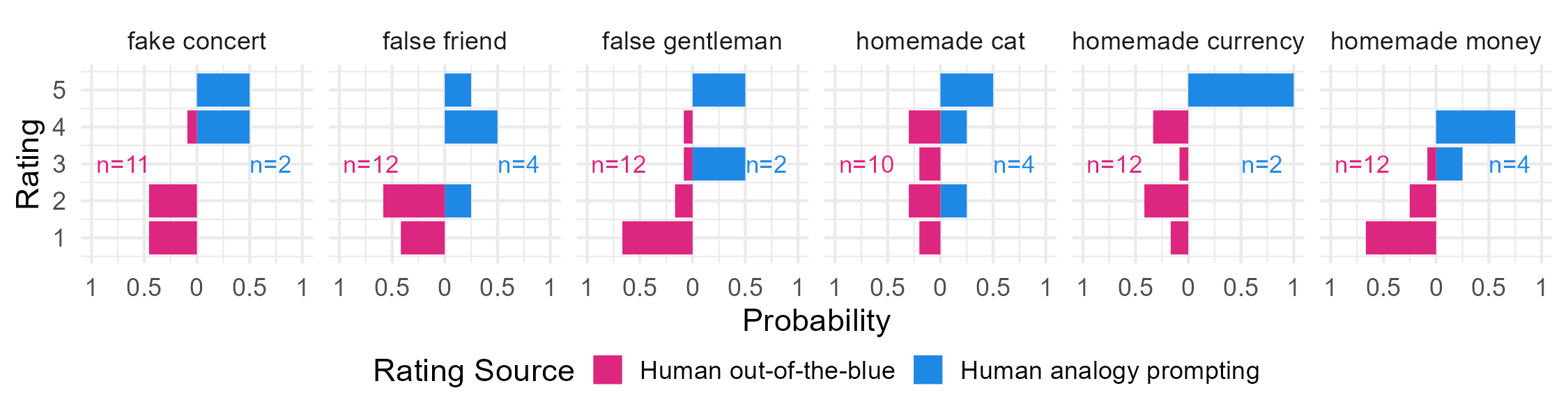}
  \caption{Distributions for the 6 bigrams with the highest JS divergences when an analogy is used. $n =$  number of ratings in each distribution; for analogy prompting, this is however many people found an analogy.}
  \label{fig:analogy-different-dists}
\end{figure*}

Overall, participants self-reported that they could find an analogy for 56.4\% of responses. For every bigram except \textit{fake impression}, at least one person was able to find an analogy, although 13 of 143 participants never produced an analogy. A plot of analogy availability for each bigram is shown in Figure \ref{fig:analogy-availability-bars} in Appendix \ref{app:analogy-availability-figure}.\footnote{We attempted a regression to predict analogy availability but found nothing of interest; see Appendix \ref{app:analogy-hard}.}

\paragraph{Type of analogy.} Figure \ref{fig:analogy-types} shows statistics for the types of analogy drawn. We find that 58.4\% of analogies use the same adjective as the original bigram, such as \textit{knockoff watch} $\rightarrow$ \textit{knockoff purse}, while only 10\% change the adjective and use the same noun, such as \textit{homemade money} $\rightarrow$ \textit{counterfeit money}. 
A further 6.2\% of analogies use a single noun. While a number of these single-noun analogies seem intended as same-adjective analogies (such as \textit{tiny bed} $\rightarrow$ \textit{(tiny) chair}), we do see some interesting cases such as \textit{artificial rumor} $\rightarrow$ \textit{lie}, which may not be an analogy in the strict sense but are still solving the task by mapping to a known phrase.
The remaining 25.4\% use a different adjective/modifier and noun.

Qualitatively, we see that our participants reach for a much wider set of concepts than our analogy model when drawing analogies; choices such as \textit{homemade lake} $\rightarrow$ \textit{homemade cookies}, \textit{false impression} $\rightarrow$ \textit{wrong interpretation} or even \textit{multicolored weapon} $\rightarrow$ \textit{painted nails} are common. Participants are more likely than our model to reach for nouns that are not that similar to the original noun but are highly associated with the adjective, such as \textit{knockoff purse} (11 occurrences as analogy), \textit{counterfeit money} (10 occurrences), \textit{homemade cookies} or \textit{illegal immigrant} (3 occurrences each). 

\paragraph{Distribution shift.} Does analogical reasoning shift the distribution compared to the original ratings gathered by \citet{ross_fake_2025}, where no instructions on how to reason were provided?
In the cases where an analogy was found, we find an average JS divergence of 0.16 overall between bigram distributions in this experiment vs.~in  \citet{ross_fake_2025}, with 0.21 on privative-type adjectives (0.32 for \textit{fake}), 0.35 on \textit{homemade} (recall that nouns for \textit{homemade} were picked adversarially to be more likely to be privative) and 0.14 on zero-frequency (presumed novel) bigrams. 

We also conduct Kolmogorov-Smirnoff tests per-bigram (with Holm-Bonferroni adjustment) to determine which of the distributions are significantly different. Since our $n$ per bigram is quite small for statistical purposes (at best $n=12$, lower if not all participants found an analogy for the bigram), no bigrams are significantly different. We cannot conclude from this that the distributions are indeed the same when analogy is used; the sample size is just too limited. 
Instead, we plot the distributions for 6 bigrams with the highest JS divergences in Figure \ref{fig:analogy-different-dists}. 
The divergence for \textit{homemade currency} and \textit{homemade money} (and to a lesser extent \textit{false friend}) is particularly striking: analogy leads people to dramatically different inferences in these cases, since most \textit{homemade} and many \textit{false} items (such as \textit{false rumor}) still clearly qualify as an instance of the noun.

\paragraph{Correlation between analogy availability and distribution shift.} We fit a beta regression in R \citep{brooks_glmmtmb_2017} 
that predicts JS divergence as a function of analogy availability. We find a strong negative correlation: JS divergence decreases 
as analogy availability increases
($p < 0.001$). In other words, the harder it is to find an analogy, the more likely any analogies that are found will lead people astray from the original distribution.

\subsection{Discussion}

This experiment shows that analogy is a viable approach for many bigrams, and in many cases results in similar judgments as in \citet{ross_fake_2025}, where participants could reason freely. 
However, for several bigrams such as \textit{homemade money}, using an analogy yields dramatically different inferences, suggesting that analogy was not used to derive the original distribution. 
We also see bigrams where people struggle to come up with any analogy at all, such as \textit{fake impression} ($n=0$). This was the case for 10 of our 35 zero-frequency bigrams ($n \leq 50\%$), putting into question the viability of analogical reasoning for generalization. 
Our analogy model also shows a higher-than-average JS divergence for all bigrams (except one) where analogical reasoning substantially shifts human ratings. It also shows a higher-than average JS divergence for over half the bigrams where humans struggle to come up with an analogy.
Overall, a linear regression predicting human JS divergence from the analogy model's JS divergence explains 40\% of variation
, suggesting that analogy serves as a viable explanation for some, but not all of the variation in human inferences.
As for LLM behavior, human analogy availability and human-human JS divergence when using analogies both correlate poorly with LLM-human JS divergence per-bigram, with $R^2 = 0.05$ in both cases ($p = 0.03$ and $p = 0.04$ respectively). A similar regression with our analogy model in Section \ref{sec:am-llms} also showed low correlation. This suggests that analogical reasoning poorly explains LLM behavior, corroborating our previous conclusion in Section \ref{sec:am-llms}. 

Finally, we observe that our participants use a much broader definition of ``analogy'' than our analogy model (or the examples we gave during training), suggesting that our model adheres to adjective and noun similarity overly strictly. Further, our analogy model is strictly non-compositional at the meaning level, whereas some human analogies such as \textit{false impression} $\rightarrow$ \textit{wrong interpretation} may well be arising from the participants first composing the meaning of \textit{false impression} and then looking for phrases with a similar meaning.\footnote{\textit{False} may mean \textit{not truthful/insincere} or just \textit{fake} (as in \textit{false teeth}); the choice of meaning depends on the noun.}

\section{Conclusion}

\citet{ross_fake_2025} claim that humans must be handling adjective-noun bigrams compositionally, since they draw consistent inferences about novel bigrams, and \citet{ross_is_2024} take LLMs' capacity to draw reasonably human-like inferences on the same novel bigrams as evidence for composition. We explored the possibility that this generalization might be explained without composition in either or both cases, specifically by analogical reasoning over adjective and nouns using previously encountered and memorized inferences. 

\paragraph{Composition in humans.}
We find that while many of the novel bigrams in the dataset can indeed be handled successfully by analogy, analogy is not sufficient to explain human behavior fully.
Our analogy model diverges significantly from human distributions on 20 bigrams and shows insufficient generalization to zero-frequency bigrams, with a JS divergence of 0.25 from humans. Humans both struggle to come up with analogies for 24\% of bigrams tested and are led astray when they do for several bigrams, such as \textit{homemade currency}.
We thus conclude that analogical reasoning is a successful strategy for generalization in a remarkable proportion of the dataset of \citet{ross_fake_2025}, but analogy does not suffice to handle the full data. Thus, their conclusion that some mechanism of composition seems necessary to handle the whole range, \textit{homemade currency} and all, is supported---even if humans need not (and judging by our data, quite possibly do not) invoke it in every case. 
This conclusion is similar to the result of \citet{albright_rules_2003}, who found that an analogical model of English past tense morphology did not explain participant behavior well, and concluded that speakers used abstract rules to generalize rather than analogy.

\paragraph{Composition in LLMs.}
We likewise find that LLM behavior can be partially, but not fully explained by analogical reasoning. 
Our analogy model is unable to reach the performance of the most successful LLMs in \citet{ross_is_2024}, in particular when generalizing to zero-frequency bigrams. 
Moreover, a linear model predicting LLM JS divergence as a function of analogy model JS divergence only explains 16\% of the variance. 
While this does not prove that Llama 3 70B Instruct is conducting \textit{bona fide} composition, it provides exciting indications that it might---at minimum, Llama 3 70B Instruct is better able to incorporate the interaction between the adjective meaning and noun meaning 
than our purely word analogy-based model. Investigating how composition, typically conceptualized as abstract rules, can be implemented in LLMs would be an interesting avenue of future research---the \textit{abstraction-via-exemplars} account discussed in \citet{misra_2023_abstraction} may provide a promising starting point.

\paragraph{Standards of evidence for composition} This paper contributes to a broader discussion about the standards of evidence required for composition \citep{mccurdy_toward_2024,pavlick_not-your-mothers-connectionism_2025}.
If behavioral experiments about generalization can provide evidence about composition (and not all researchers believe they can), we must be sure to rule out other methods of generalization such as analogy. We further need to ensure we have a precise enough definition of compositionality to capture our intuition that analogy, by virtue of referring to information not (obviously) included in the meanings of the parts, is not a kind of composition \citep{szabo_case_2012}. By making an explicit model of analogical reasoning, we can both show the way in which it requires this additional information and show that analogical reasoning fails to generalize in the expected way, relative to our human data.

\section*{Acknowledgments}

Icons in Figure \ref{fig:model-diagram} courtesy of The Noun Project, from artists Atacan, Goat\_Team, Ricons, vecto and Zaenal Abidin.

\bibliography{custom}

\appendix

\section{Analogy Availability for Humans} \label{app:analogy-availability-figure}

\begin{figure*}[t]
\centering
    \includegraphics[width=\linewidth]{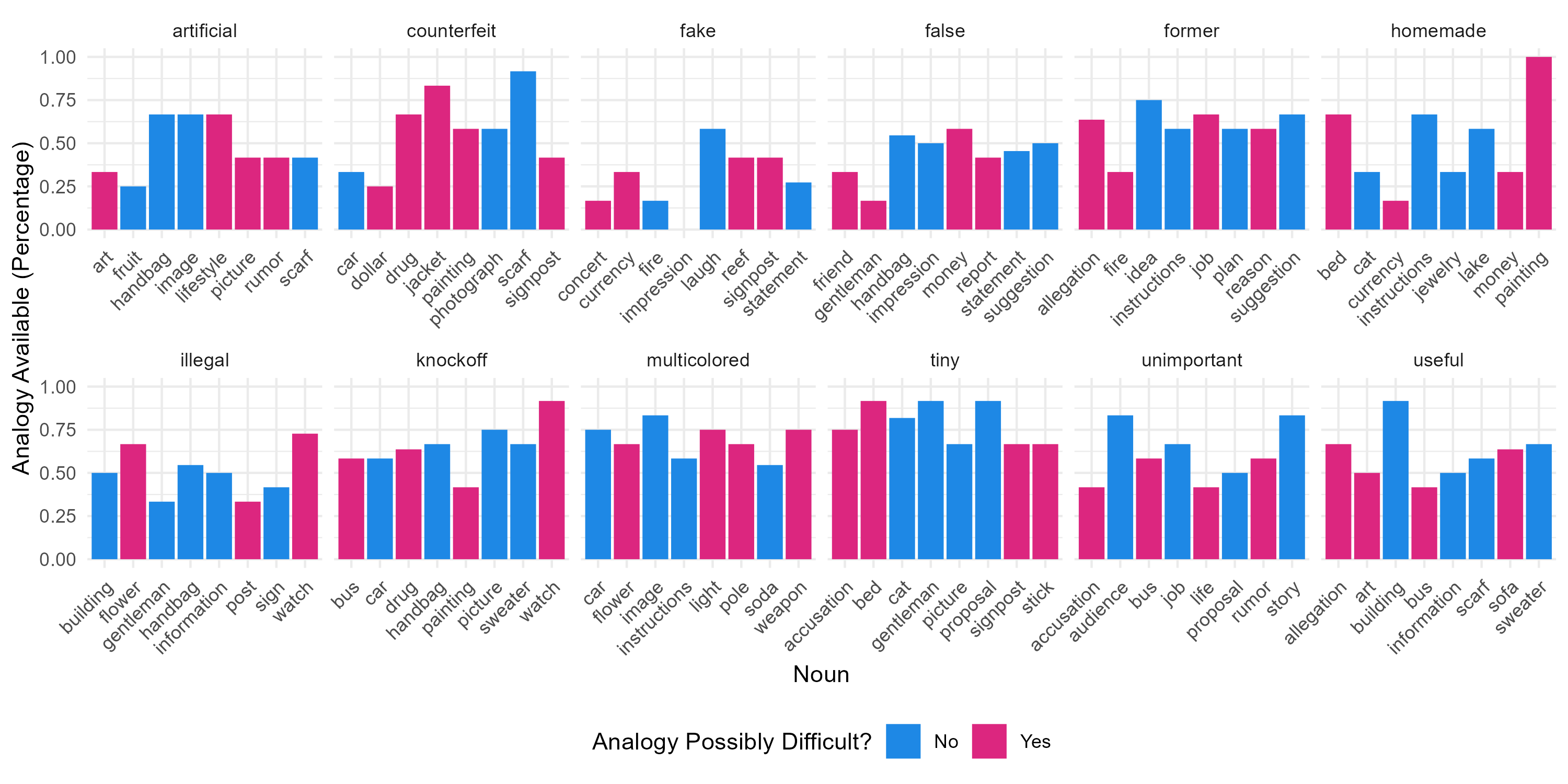}
  \caption{Analogy availability for all 96 bigrams in the analogy prompting experiment.  Color indicates whether it was predicted in advance that it might be difficult to find an analogy, based on the ratings from \citet{ross_fake_2025} in conjunction with noun frequencies and WordNet-based distance measures (see Section \ref{app:analogy-hard}).}
  \label{fig:analogy-availability-bars}
\end{figure*}

Figure \ref{fig:analogy-availability-bars} shows the percentage of times participants were able to find an analogy for each bigram, colored by the estimation of analogy difficulty discussed in Appendix \ref{app:analogy-hard}.

\section{Participant Recruitment Criteria} \label{app:native-speaker}

For our experiment in Section \ref{sec:human-experiment}, we recruit people on Prolific who self-report English as their first and primary language and are located in the US. We further ask them at the end of the study whether they
learned English before the age of 5 and whether they speak American English---if not, they are paid but excluded from the analysis.
This implementation of ``native speaker'' is merely intended as a practical way to expect shared language experiences among our participant sample \citep{cheng_problematic_2021}.

\section{Detailed Analogy Model Results} \label{sec:hyperparameters}

\subsection{Model Configuration}

As discussed in Section \ref{sec:analogy-model-algorithm}, the model has three configurable parameters: whether to do analogy over just nouns or also to include up to one adjective (``Noun only'' vs.``Noun + Adjective''), how many nearby bigrams to retain ($k$), and whether to return the memorized distributions from the training set when asked about a bigram in the training set, or to apply the algorithm as if that particular bigram were not known.

We consider only up to 1 adjective since a hyperparameter search over up to 10 adjectives showed that 1-2 adjectives were always optimal; moreover, we only have 12 candidate adjectives to begin with, and manual inspection suggests that at most 1-2 of them ought to be relevant.

We consider 100 nearby nouns since we do not want to artificially constrain our model and prevent it from finding enough bigrams that it actually knows.
Having separate steps for adjective/noun retrieval, assembling candidate bigrams, and then checking which bigrams are known is an artificial implementation choice that we make for our algorithm; humans could well be retrieving similar nouns and checking whether the resulting bigram is known in tandem. Thus,
we always retrieve 100 nearby nouns ``just in case'' and instead rely on the number of bigrams $k$ to constrain the model. 
As discussed in Section \ref{sec:analogy-model-algorithm}, we set $k \leq 5$ to impose constraints akin to human working memory \citep{cowan_magical_2001,adam_clear_2017}. 
We allow the model to do a grid search over the exact value of $1 \leq k \leq 5$ by evaluating the model on the training set with memorization disabled. The optimal $k$ typically ranges between 3-5 bigrams.
In Table \ref{tab:analogy-model-results}, we also report the special configuration $k=1$, where the model only considers the most similar bigram it can come up with. This mimics humans going with the ``first bigram they can come up with'', assuming that their retrieval process chooses a good candidate as its first choice. 

The final configuration choice, which we did not discuss in Section \ref{sec:analogy-model-algorithm}, is the training data -- what should be considered as bigrams that humans have previously encountered.
Option 1 is to include all bigrams classed as ``high frequency'' by \citet{ross_is_2024}, i.e. all bigrams in the top quartile of their dataset.
This results in sparse data for some adjectives. Notably, this only includes a single bigram involving the adjective \textit{knockoff} 
and no bigrams including \textit{unimportant},
meaning the model will be at a disadvantage for bigrams with these adjectives. In the N+A setting, it will have to rely primarily on bigrams involving e.g. \textit{counterfeit}; in the noun only setting, it will often return no distribution. It is unclear whether this sparsity is precisely realistic, because these adjectives and their bigrams are low-frequency, or not. 
Options 2a and 2b are to train on the top $x$ most frequent bigrams for each adjective, where we can consider (a) $x=5$ (akin to the $k \leq 5$ setting for nearby bigrams), or (b) $x=23$, which results in a nearly identical size training set (276 bigrams) to taking the top quartile (279 bigrams). We report all three settings in Table~\ref{tab:analogy-model-results}.

Finally, in the case where no similar bigrams have known ratings, we opt to return a null distribution, which is always incorrect. We could alternatively return a fallback distribution which concentrates all its probability mass on ``Unsure'', but this will also be very unlike the human distributions under the Jensen-Shannon metric (which tend to have high SD when not concentrated at the ends of the scale), so this makes little difference. 
In practice, this only occurs in the ``Noun only'' setting for some bigrams involving \textit{knockoff} and \textit{unimportant} when we use the top quartile of bigrams as the training set, since these adjectives have few or no high-frequency bigrams (1 for \textit{knockoff}, 0 for \textit{unimportant}).

\subsection{Detailed Results}

\begin{table*}[ht]
    \centering
    \begin{tabular}{lccccc}
    \toprule
    & \multicolumn{5}{c}{JS Divergence (lower is better)} \\
    Model & Novel bigr. & Zero-freq. bigr. & Privative A & Total & Total (+\texttt{mem}) \\ \midrule
    Human (resampled) & N/A & 0.04 & 0.05 & 0.04 & N/A \\
    Human (analogy exp.) & N/A & 0.14 & 0.21 & 0.16 & N/A \\
    Llama 3 70B Instruct & N/A & 0.17 & 0.26 & 0.17 & N/A \\
    Uniform distr. baseline & N/A & 0.33 & 0.20 & 0.34 & N/A \\
    \midrule
    \textbf{Analogy models: GloVe} & & & & & \\
    N only, $k=1$, top qt. & 0.44 & 0.57 & 0.45 & 0.39 & 0.29 \\
    N only, $k=1$, top 5/A & 0.32 & 0.34 & 0.44 & 0.32 & 0.30 \\
    N only, $k=5$, top qt. & 0.41 & 0.55 & 0.39 & 0.36 & 0.27 \\
    N only, $k=3$, top 5/A & 0.28 & 0.28 & 0.36 & 0.28 & 0.25 \\
    N only, $k=4$, top 23/A & \underline{0.26} & \underline{0.25} & \underline{0.33} & \underline{0.26} & \underline{0.17} \\
    N+A, $k=1$, top qt. & 0.29 & 0.31 & 0.39 & 0.29 & 0.19  \\
    N+A, $k=4$, top qt. & \underline{0.26} & 0.26 & 0.34 & \underline{0.26} & \underline{0.17} \\
    N+A, $k=3$, top 5/A & 0.27 & 0.27 & 0.36 & 0.27 & 0.25 \\
    N+A, $k=3$, top 23/A & \textbf{0.25} & \underline{0.25} & \textbf{0.32} & \underline{0.26} & \underline{0.17} \\
    \textbf{Analogy models: WordNet} & & & & & \\
    N only, $k=1$*, top qt. & 0.41 & 0.54 & 0.36 & 0.36  & 0.26 \\
    N only, $k=1$*, top 23/A & \textbf{0.25}  & \textbf{0.24} & \textbf{0.32} & \textbf{0.25}  & \textbf{0.16}  \\
    \multicolumn{6}{l}{\textbf{Analogy models: Llama 3 70B embeddings (final layer)}}  \\
    N only, $k=1$, top qt. &  0.44  & 0.53  & 0.44 & 0.40 & 0.28 \\
    N only, $k=4$, top qt. &  0.40  &  0.50 & 0.37  &  0.35  &  0.26 \\
    N only, $k=5$, top 23/A & \underline{0.26} & 0.26 & 0.34  &  \underline{0.26} &  \underline{0.17} \\
    N+A, $k=1$, top qt. & 0.33 & 0.33 & 0.44 & 0.34 & 0.22 \\
    N+A, $k=4$, top qt. & 0.28 & 0.27 & 0.35 & 0.28 & 0.18 \\
    N+A, $k=5$, top 23/A & 0.27 & 0.26 & 0.34 & 0.28 & 0.18 \\
    \multicolumn{6}{l}{\textbf{Analogy models: Llama 3 70B embeddings (initial layer)}}  \\
    N+A, $k=5$, top qt. & 0.28 & 0.30 & 0.35 & 0.27 & 0.18 \\
    \bottomrule
    \end{tabular}
    \caption{Average JS divergence (\textbf{best} / \underline{second}) between various configurations of analogy models and human rating distributions, with \& without training data memorization, for `N only' vs. `N+A' (1 nearby adjective) and $k = 1$ vs.~$k \leq 5$ nearby bigrams (exact value of $k$ tuned on training data). `Novel bigrams' = bigrams held out from each analogy model -- for humans and LLMs, we can only be sure that zero-frequency bigrams are novel. `Privative A' = bigrams with ``privative'' adjectives. * = set $k \leq 5$ but tuning chose $k=1$. Llama 3 results and baseline from \citet{ross_is_2024}. 
    }
    \label{tab:analogy-model-results}
\end{table*}

Table \ref{tab:analogy-model-results} shows the results for the analogy models built with GloVe embeddings, comparing the noun only setting with the N+A (noun + adjective) setting, and the single bigram setting ($k=1$) with $k \leq 5$. We report the exact value for $k$ chosen by the hyperparameter search. We also compare training on the top quartile of bigrams vs. training on the top 5 or 23 per adjective. Note that for the top 5 case, the set of novel bigrams (column 2, ``Novel bigr.'') is larger than in the other cases. 
We find that the simplest setting, analogy to a single noun (N only, $k=1$) does not outperform a uniform distribution baseline overall. 
However, if we allow multiple adjectives, analogy to a single bigram ($k=1$) is sometimes the best (selected even when we tune on $k \leq 5$). We also achieve similarly good results if we use nouns only but allow averaging over $k \leq 5$ bigrams.
In the noun + adjective case, results are also similar whether we train on the top quartile of bigrams or the top 23 bigrams per adjective -- training set size appears to be the driving factor, not how it is balanced. However, in the noun only case, which includes all the WordNet models, we unsurprisingly see a performance boost from including more bigrams for each adjective. (When training on the top quartile, the noun only setting necessarily fails for all bigrams involving \textit{unimportant}, since there is no bigram with \textit{unimportant} in the training data, and does poorly for \textit{knockoff} as well, since there is only one bigram with \textit{knockoff} in the training set.)
Memorization of the training set boosts overall performance, as expected, though not so much when the training set is very small (top 5 bigrams per adjective). 

Further, we observe that performance is generally lower on privative adjectives than overall, which makes sense because many bigrams with subsective adjectives have distributions almost entirely consolidated around ``Definitely yes'', and can be predicted from other bigrams. 

\subsection{Significantly Different Distributions} \label{sec:am-signif-different}

\begin{figure*}[t]
  \includegraphics[width=\textwidth]{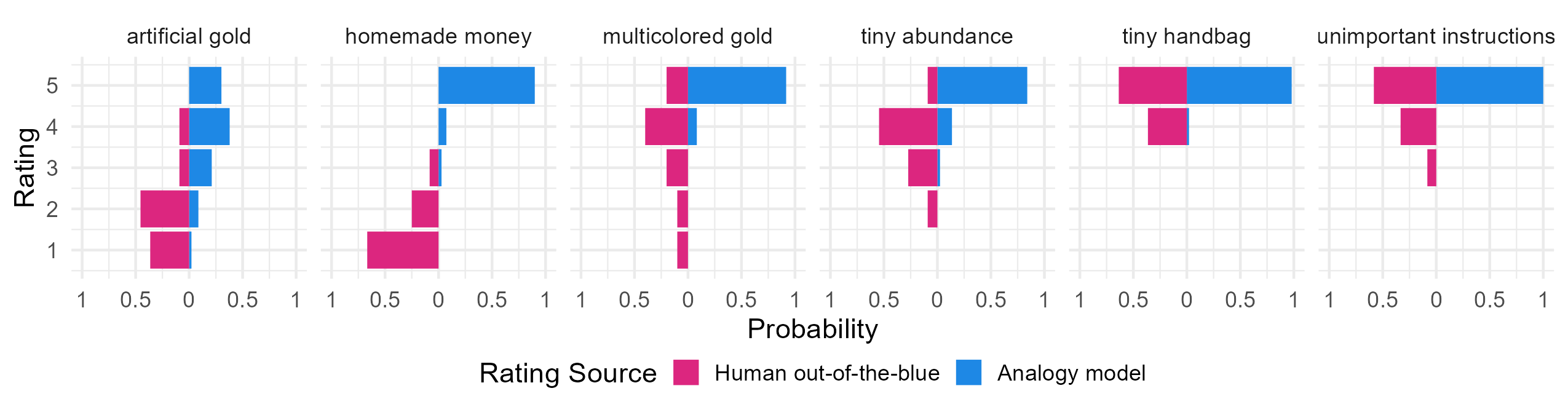}
  \caption{Difference between distributions for 6 of the 10 bigrams which are significantly different between the analogy model (even with \texttt{mem}) and the original human distributions. In each case, the model predicts more subsective ratings than humans.}
  \label{fig:analogy-model-different-dists}
\end{figure*}

Figure \ref{fig:analogy-model-different-dists} shows 6 of the 10 bigrams where the analogy model (GloVe, $k \leq 5$, with \texttt{mem}) predicts a significantly different distribution according to the Kolmogorov-Smirnoff test (with Holm-Bonferroni adjustment) in Section \ref{sec:am-human-discussion}.

\section{Estimate of Analogy Difficulty} \label{app:analogy-hard}

\subsection{Overview}

For our analogical reasoning experiment, 
we attempt to estimate which bigrams might be difficult to find analogies for and balance evenly for this. 
We suppose that analogy could be difficult for bigrams with one or more following qualities: 

\begin{itemize}[nosep]
    \item the noun has no high-frequency neighbors (below median among the nouns in the dataset)
    \item there are multiple convergent nearby bigrams with ratings that conflict
    \item there are non-convergent nearby bigrams (i.e. bigrams for which the conclusion is uncertain)
\end{itemize}

We use WordNet \citep{miller_wordnet_1995} rather than word embeddings to find neighboring nouns, since WordNet is manually annotated by human experts, and the British National Corpus for noun frequencies \citep{leech_word_2014}. We manually define adjective similarity, since WordNet only provides a hierarchical taxonomy -- and thus, a similarity metric -- over nouns, described in Section \ref{sec:adj-similarity}. 

\subsection{Results}

In fact, we find that these criteria do not predict how often participants were able to come up with an analogy. 

We fit a logistic mixed effects model in R \citep{bates_fitting_2015} 
that predicts whether participants could find an analogy or not. As fixed effects, we include the three factors described in Section \ref{sec:human-exp-method}, as well as adjective class (typically privative or subjective) and specificity of the noun (measured by depth in the Wordnet taxonomy). We include adjective and participant as random effects. We also fit a second model where we replace specificity of noun with bigram frequency (the two are too correlated to include in the same model). 
In fact, we find that none of these factors are significant ($p < 0.05$) except for the presence of nearby divergent bigrams. This feature, however, only applies to 6 bigrams in the experiment, so this may just be spurious.
This non-significance may be the result of many false negatives in our labeling of these factors, since we can only test for nearby bigrams among the bigrams that \citet{ross_fake_2025} studied, not among the totality of nearby bigrams. It may also result from our participants construing analogy much more broadly than we did, as discussed in Section \ref{sec:human-exp-results}.

\subsection{Adjective Similarity Details} \label{sec:adj-similarity}

We use the following (asymmetric) similarities, which are approximately scaled to match the Wu-Palmer similarity metric (which is 0.5 for siblings). 

\begin{enumerate}[nosep]
    \item \textit{artificial} $\rightarrow$  \textit{fake, false}: 0.75 \\ \phantom{\textit{artificial}} $\rightarrow$ \textit{counterfeit, knockoff}: 0.5
    \item \textit{counterfeit} $\rightarrow$  \textit{knockoff}: 0.9 \\ \phantom{\textit{counterfeit}} $\rightarrow$ \textit{fake, false}: 0.75 \\ \phantom{\textit{counterfeit}} $\rightarrow$ \textit{artificial}: 0.5 
    \item \textit{fake} $\rightarrow$ \textit{artificial, counterfeit, false, } \\ 
    \phantom{\textit{fake} $\rightarrow$} \textit{knockoff}: 0.75
    \item \textit{false} $\rightarrow$ \textit{fake}: 0.9 \\ \phantom{\textit{false}} $\rightarrow$ \textit{counterfeit, knockoff, artificial}: 0.75 
    \item \textit{knockoff} $\rightarrow$ \textit{counterfeit}: 0.9 \\ \phantom{\textit{knockoff}} $\rightarrow$ \textit{fake}: 0.75 
    \item \textit{former} $\rightarrow$ \textit{artificial, counterfeit, fake, } \\ \phantom{\textit{former} $\rightarrow$} \textit{false, knockoff}: 0.5 
    \item \textit{homemade} $\rightarrow$ \textit{artificial, fake, false}: 0.8 \\ \phantom{\textit{homemade}} $\rightarrow$ \textit{tiny, multicolored}: 0.75 \\ \phantom{\textit{homemade}} $\rightarrow$ \textit{useful, illegal,} \\ \phantom{\textit{homemade} $\rightarrow$} \textit{unimportant}: 0.5 
    \item The remaining 5 subsective adjectives, \textit{useful, tiny, illegal, unimportant} and \textit{multicolored} are all assigned a similarity of 0.5 to each other and to \textit{homemade}.
\end{enumerate}

Note that we provide an unusually privative-looking set of similarities for \textit{homemade} since the examples with \textit{homemade} in the experiment are disproportionately chosen to be less subsective and thus challenging for analogy.
Moreover, these similarities are adjusted for the fact that these are the only 12 adjectives available -- of course they would be scaled differently if there were more options.
We do not expect small changes to these similarities to have a noticeable difference on the selected bigrams.

\section{Using Human Analogy Bigrams in the Analogy Model} \label{sec:am-extra-data}

One bottleneck of our analogy model appears to be its lack of available bigrams with which to draw an analogy, i.e.~which it has ratings for, compared to humans. We can try to ameliorate this by additionally giving it all the analogies found in the human analogy experiment, by assuming that the rating that they provide for the target bigram is the same as the rating they would assign to the analogical bigram. (This should be true if they are using the analogy as intended.) 
We filter the provided analogy phrases through WordNet to retain only two-word phrases whose first word is an adjective and the second a noun. This adds 340 bigrams involving 91 adjectives and 260 nouns. (The original dataset contained only 12 adjectives and 102 nouns.)

Unfortunately, we do not have full distributions for these bigrams; only 68 of the 340 bigrams so found have more than one rating, and only 11 have more than three. For target bigrams with privative adjectives, whose distributions are often spread out, analogy to these new bigrams will thus yield a high JS divergence simply because the distribution is too sparse. 
In line with this, the results in Table~\ref{tab:analogy-model-results-ap} show that adding these additional bigrams worsens or does not improve the two best-performing GloVe models from Table~\ref{tab:analogy-model-results}, though it does result in different hyperparameter choices during the grid search ($k \leq 5$). 

To compensate for only having single ratings, we can instead evaluate the analogy models 
with the more lenient ``accuracy within 1 SD of the human mean'' metric proposed for single ratings by \citet{ross_is_2024}, which lets the model predict a mean rating instead of a full distribution. It is then judged ``accurate'' (enough) if this rating falls within 1 SD of the mean of the human rating distribution that bigram (rounded to the nearest integer), incorrect otherwise. 
The problem with this metric, besides being ad-hoc, is that the simple ``majority'' baseline described in \citet{ross_is_2024}, which simply guesses ``Unsure'' for all bigrams with privative adjectives and ``Definitely yes'' for all those with subsective adjectives, achieves an accuracy of 0.89 using this metric. 
Bigrams with privative adjectives generally have such a high SD that this is a large and easy target to hit. 
Nonetheless, a random guessing baseline scores only 0.46 on this metric, so the metric is still somewhat informative. 
 
If we add the new bigrams provided by the analogy prompting experiment to the training set and evaluate with this Within 1 SD metric, we do see
a significant performance increase compared to using just the original training set, as shown in Table~\ref{tab:analogy-model-results-ap-within-1-sd}. 
Note that optimizing over this metric yields new values for the parameter $k$, within the constraint $k \leq 5$. $k=1$ is uniformly chosen during tuning even when we set $k \leq 5$. 
In contrast to the JS divergence, where we generally saw lower (better) values for subsective adjectives and higher (worse) values for privative ones, this metric yields the opposite, since the SDs for subsective-adjective bigrams are much smaller: we see lower (worse) accuracies for subsective adjectives.

This suggests that if we had full distributions for these bigrams, adding more training data might indeed significantly improve the model. What amount of training data is appropriate for modeling humans remains an open question. 

\begin{table*}[t]
    \centering
    \begin{tabular}{lccccc}
    \toprule
    & \multicolumn{5}{c}{JS Divergence (lower is better)} \\
    Model & Novel bigr. & Zero-freq. B & Privative A & Total & Total (+\texttt{mem}) \\ \midrule
    N+A, $k=4$, top qt. & 0.26 & 0.26 & 0.34 & \textbf{0.26} & \textbf{0.17} \\
    N+A, $k=4$, top qt.~+ exp. & 0.45 & 0.62 & 0.41 & 0.39 & 0.29 \\
    N+A, $k=3$, top 23/A & \textbf{0.25} & \textbf{0.25} & \textbf{0.32} & \textbf{0.26} & \textbf{0.17} \\
    N+A, $k=4$, top 23/A + exp. & 0.26 & 0.26 & 0.33 & \textbf{0.26} & \textbf{0.17} \\
    \bottomrule
    \end{tabular}
    \caption{Average JS divergence (\textbf{best}) between analogy models and human rating distributions for the best GloVe models in Table \ref{tab:analogy-model-results} and their counterparts trained on the additional bigrams from the human analogy experiment. 
    This additional training data does not improve model performance as measured by JS divergence, because we do not have full distributions for many of the additional bigrams. 
    }
    \label{tab:analogy-model-results-ap}
\end{table*}

\begin{table*}[t]
    \centering
    \begin{tabular}{lccccc}
    \toprule
    & \multicolumn{5}{c}{Accuracy within 1 SD of human mean} \\
    Model & Novel bigr. & Zero-freq. B & Privative A & Total & Total (+\texttt{mem}) \\ 
    \midrule
    ``Majority'' baseline & N/A & 0.91 & 0.78 & 0.89 & N/A  \\
    Random guessing baseline & N/A & 0.46 & 0.61 & 0.46 & N/A  \\
    \midrule
    N+A, top qt. & 0.71 & 0.77 & \textbf{0.72} & 0.69 & 0.78 \\
    N+A, top qt.~+ exp. & \textbf{0.76} & 0.76 & 0.69 & \textbf{0.74} & \textbf{0.81} \\
    N+A, top 23/A & 0.70 & 0.76  & 0.71 & 0.68  & 0.76 \\
    N+A, top 23/A + exp. & 0.75 & \textbf{0.79} & \textbf{0.72} & \textbf{0.74} & 0.80 \\
    \bottomrule
    \end{tabular}
    \caption{Results for the best GloVe models in Table \ref{tab:analogy-model-results} and their counterparts trained on the additional bigrams from the human analogy experiment using the more lenient ``accuracy within 1 SD of human mean'' metric proposed by \citep{ross_is_2024}. 
    All models use $k=1$ even when tuned with $k \leq 5$; this makes sense as averaging is less likely to improve this metric.
    Unlike for the JS divergence shown in Table \ref{tab:analogy-model-results-ap}, results do improve.
    However, results must be interpreted relative to the ``majority'' baseline provided by \citep{ross_is_2024}, which highlight the difficulty with this metric. 
    }
    \label{tab:analogy-model-results-ap-within-1-sd}
\end{table*}

\section{Experiment Training Instructions} \label{app:experiment-training}

The instructions provided to participants are shown in Table \ref{tab:exp-instructions}.

\begin{table*}
\centering
    \begin{tabular}{p{0.97\textwidth}}
    \toprule 
    This survey involves questions of the form ``Is a toy hippo still large?'' We're interested in whether it's possible to solve these kinds of questions by reasoning using a similar phrase that you already know the answer for (``by analogy''), such as ``toy hippo'' $\rightarrow$ ``toy elephant'' (toy elephants are usually not large). For the purposes of this survey, the similar phrase / analogy can be another similar thing, or a class of things (like animals or gadgets). The important part is that you know the answer for the new phrase without having to think about it. \\
    \\
    Let's start with three examples that demonstrate how the survey works and what we mean by analogy. \\
    \\
Each question consists of two parts. First you will answer whether you can think of a suitable analogy (yes/no), and type in the similar phrase if you answered yes. The phrase should consist of 1-3 words and will typically be of the form "[adjective] [noun]". Then you will attempt to answer the original question (e.g. "Is a toy hippo still large?") using the phrase you chose, or without it if you couldn't think of one. \\
\\
Please pay close attention to the following examples, as we will ask you to follow this style of reasoning in the rest of the survey. \\
\midrule
Is melted plastic still plastic? \\
\\
\textit{Can you think of an analogy to another similar phrase that would help answer this question?} \\
\\
You can think of an analogy from ``melted plastic'' $\rightarrow$ ``melted wax'' or ``melted chocolate.'' This is useful because you immediately know the answer to ``Is melted wax still wax?'' or ``Is melted chocolate still chocolate?'' So, you would answer ``yes'' to this question and type ``melted wax'' or ``melted chocolate'' in the text box below. \\
\\
\textit{Based on the analogy you chose: }\\
Is melted plastic still plastic? \\
\\ 
Because melted wax is still wax (or melted chocolate is still chocolate), you conclude that melted plastic is still plastic, or probably still plastic. So, you would answer ``Definitely yes'' or ``Probably yes'' depending on your interpretation. \\
\midrule 
Is a hard-boiled egg still runny? \\
\\
\textit{Can you think of an analogy to another phrase that would help answer this question?} \\
\\
You probably find it hard to quickly think of an analogy that can help answer the question. While you may be able to come up with similar phrases, they don't immediately provide an obvious answer. So, you would answer ``No'' to this question. \\
\textit{[Instructions for second part irrelevant, omitted]} \\
\midrule
Is a decorative pumpkin still edible? \\
\\
\textit{Can you think of an analogy to another similar phrase that would help answer this question?} \\
\\
As in the previous example, it is hard to quickly think of an analogy that can help answer the question. While you may be able to come up with similar phrases, they don't immediately provide an obvious answer. So, you would answer ``No'' to this question. \\
\textit{[Instructions for second part irrelevant, omitted]} \\
\bottomrule
    \end{tabular}
    \caption{Training instructions and examples shown to participants to demonstrate what we intend by ``analogy''.}
    \label{tab:exp-instructions}
\end{table*}

\end{document}